\documentclass[11pt]{article}
\usepackage{paclic29}
\usepackage{times}
\usepackage{latexsym}
\usepackage{amsmath}
\usepackage{multirow}
\usepackage{url}
\usepackage{graphicx}
\usepackage{epsfig}
\usepackage{graphicx,dblfloatfix}
\usepackage{tabularx} % 可以设置表格宽度
\usepackage{booktabs} % 三线表加强
\usepackage{algorithm} 
\usepackage{algorithmic}
\usepackage[hang,nooneline]{subfigure}

\title{ An Empirical Study on Sentiment Classification of Chinese Review
using Word Embedding }
\author{Yiou Lin \qquad Hang Lei \qquad Jia Wu \qquad Xiaoyu Li \\School of Information and Software Engineering \\
 University of Electronic Science and  Technology of China\\Chengdu, China \\
  {\tt lyoshiwo@gmail.com \{hlei,jiawu, xiaoyuuestc\}@uestc.edu.cn}\\
 }
  
\date{}

\begin{document}
\maketitle
\begin{abstract}
  In this article, how word embeddings can be used as features in Chinese sentiment classification is presented. Firstly, a Chinese opinion corpus is built with a million comments from hotel review websites. Then the word embeddings which represent each comment are used as input in different machine learning methods for sentiment classification, including SVM, Logistic Regression, Convolutional Neural Network (CNN) and ensemble methods. These methods get better performance compared with N-gram models using Naive Bayes (NB) and Maximum Entropy (ME). Finally, a combination of machine learning methods is proposed which presents an outstanding performance in precision, recall and F1 score. After selecting the most useful methods to construct the combinational model and testing over the 
corpus, the final F1 score is 0.920.  
\end{abstract}

\section{Introduction}
Sentiment analysis or opinion mining is the computational study of people's opinions, appraisals, attitudes, and emotions toward entities, individuals, issues, events, topics and their attributes \cite{liu2012survey}. The task of sentiment analysis is technically challenging and practically very useful. For example, businesses always want to find public or consumer opinions about their products and services. Consumers also need a sounding board rather than thinking alone while making decisions. With the development of Internet, opinionated texts from social media (e.g., reviews, blogs and micro-blogs) are used frequently for decision making, which makes automated 
sentiment analysis techniques more and more important. Among those tasks of the sentiment analysis, the key one is to classify the polarity of given texts. Many works have been done in recent years to improve English sentiment polarity classification. There are two categories of such works. One is called ``machine learning" which is firstly proposed to determine whether a review is positive or negative by using three machine learning methods, including NB, ME and SVM \cite{pang2002thumbs}. The other category called ``semantic orientation" is applied to classify words into various classes by giving a score to each word to evaluate the strength of sentiment. And an overall score is calculated to assign the review to a specific class \cite{turney2002thumbs}. 

Recently, researchers have tried to handle tasks of Natural Language Processing (NLP) with the help of deep learning approaches. Among those approaches, a useful one called word2vec has attracted increasing interest. Word2vec translates words to vector representations (called word embeddings) efficiently by using skip-gram algorithm \cite{mikolov2013distributed}. It is also proposed that the induced vector representations capture meaningful syntactic and semantic regularities, for example, ``King" - ``Man" + ``Woman" results in a vector very close to ``Queen" \cite{mikolov2013linguistic}. 

Besides, with the advancement of information technology, for the first time in Chinese history, a huge volume of Chinese opinionated data recorded in digital form is ready for analysis. Though Chinese language plays an important role in economic globalization, there are few works have been done for Chinese sentiment analysis with huge databases. It inspires us to make an empirical study on Chinese sentiment with bigger databases than usual.  

The remain of the article is organized as follows: Section 2 briefly describes related work. Section 3 describes details of the methods used in training procedure. Section 4 reports and discusses the results. Finally, we summarize our works in Section 5. 

\section{Related work}
According to \newcite{liu2012survey}, the sentiment analysis research mainly started from early 2000 by \newcite{turney2002thumbs} and \newcite{pang2002thumbs}. \newcite{turney2002thumbs} firstly used a few semantically words (e.g., excellent and poor) to label other phrases with the hit counts by queries through search engines. Then, researchers had also proposed several custom techniques specifically for sentiment classification, e.g., the score function based on words in positive and negative reviews \cite{dave2003mining} and feature weighting schemes used to enhance classification accuracy \cite{paltoglou2010study}.
Besides, the other situation of sentiment analysis is to represent texts by vectors which indicate these words appear in the text but do not preserve word order. And a machine learning approach will be used for classification in the end. In such way, \newcite{pang2002thumbs} considered classifying documents according to standard machine learning techniques. In addition, subsequent research used more features in learning, making the main task of sentiment classification engineer an effective set of features \cite{pang2008opinion}.

However, compared to English sentiment analysis, there are relatively few investigations conducted on Chinese sentiment classification until 2005 \cite{ye2005sentiment}. \newcite{li2007experimental} presented a study on comparison of different machine learning approaches under different text representation schemes and feature weighting schemes. They found that SVM achieved the best performance. After that, \newcite{tan2008empirical} found 6,000 or bigger for the size of features would be sufficient for Chinese sentiment analysis, and sentiment classifiers were severely dependent on domains or topics.  

Nowadays, inspired by the availability of large text corpus and the success of deep learning approaches, some researchers (e.g., \newcite{collobert2011natural}, \newcite{johnson2014effective}) deviated from traditional methods and tried to train neural networks such as Convolutional Neural Networks (CNN)  for NLP tasks (e.g., named entity recognition and sentiment analysis). Among them, \newcite{xu2013convolutional} and \newcite{kalchbrenner2014convolutional} got some state-of-the-art performance. But the work of~\newcite{collobert2011natural} was paid most attention for describing a unified architecture for NLP tasks which learned features by training a deep neural network even when being given very limited prior knowledge. These NLP tasks included part-of-speech tagging, chunking, named-entity recognition, language model learning and semantic role labeling.

\section{Methodology}
This section presents the methodology used in our experiment.

\subsection{Feature selection methods}
\subsubsection{Sentiment lexicon and CHI}
A sentiment lexicon accommodating sentiment words plays an important role in sentiment 
analysis. A combination of two Chinese sentiment lexicons (Hownet \cite{dong2006hownet} and DLLEX \cite{xu2008construction}) is constructed, including 30406 words in total. After removing those words which do not appear in the corpus, 10444 sentiment words are preserved. After several experiments, CHI \cite{galavotti2000feature} is chosen for information gain. Finally, 150 most valuable words are added into the new lexicon. At last, 10543 words are obtained as features.
\subsubsection{Word2vec}
Word2vec \cite{mikolov2013distributed} has gained kinds of traction today. As the name shows, it translates words to vectors called word embeddings. That is to say, it gets the vector representations of words. Gensim\footnote{http://radimrehurek.com/gensim/}, a python tool is used to get word2vec module. The method of training word2vec model is unsupervised learning and 300 is set as the quantity of the dimension of vectors. Table 1 shows the word embeddings of a Chinese hotel review which means the room is very clean and neat. For convenient display, each value of dimension is multiplied by 10,000 and indicated by $d_i$ ($ i=1,...,300$).
\begin{table}[!hbp]
\begin{tabular}{lccccc}
\hline
word & $d_1$ & $d_2$ & $d_2$ &...& $d_{300}$ \\
\hline
房间（The room）& -1102 & -202 & -668 &...& -646 \\
非常（very）& -6 & 355 & -605 &...& -460 \\
干净（clean） & -287 & -343 & 1077 &...& -232 \\
整齐（neat） & -101 & -399 & -274 &...& -986 \\
\hline
average value &-374&-148&-118&...&-581\\
\hline
\end{tabular}

\caption{An example of review vector}
\end{table}
\subsection{Traditional methods}
\subsubsection{Naive Bayes Classification}
Naive Bayes (NB) is widely used in sentiment classification which is used to classify a given review 
document $d$ to the class $c^{*}=argmax_cP(c|d)$. According to Bayes's rule,
 \[P({c_j}|d) = \frac{{P({c_j})P(d|c_j)}}{{P(d)}}\] where $c_j$ is a kind of class and $P(d)$ plays no role in selecting $c^*$. Let's mark {${f_1,f_2,...f_m}$} as the set of features that appear in all reviews, and set $n_i(d)$ as the number of times $f_i$ appears in $d$. Usually, $n_i(d)$ is set as 1, if $f_i$ appears more than one time.
Then, a formulation can be gotten as
\[P({c_j}|d) = \frac{{P({c_j})\prod\nolimits_i^m {P{{({f_i}|{c_j})}^{{n_i}(d)}}} }}{{P(d)}}\]
where the estimation of $P{({f_i}|c_j)}$ is calculated as follows, using add-one smoothing
\[\hat P({f_i}|{c_j}) = \frac{{1 + {n_{ij}}}}{{m + \sum\nolimits_{k = 1}^m {{n_{kj}}} }}\]
\subsubsection{Maximum Entropy Classification}
Maximum Entropy Classification follows the principle of maximum entropy \cite{jaynes1957information}, which means, subject to precisely stated prior data (such as a proposition that expresses testable information), the probability distribution which best represents the current state of knowledge is the one with largest entropy. 
Thus, the estimate of $P(c_j|d)$ is showed as follows
\[P\left( {{c_j}|d} \right) = {\rm{ }}\frac{1}{{\pi \left( d \right)}}{\rm{ }}exp\left( {\sum\nolimits_{i = 1}^m {{\lambda _{i,c_j}}{F_{i,c_j}}(d,c_j)} } \right)\]
\[{F_{i,c_j}}(d,x) = \left\{ {\begin{array}{*{20}{l}}
1&{if\;{n_i} > 0\;and\;x = c_j\;}\\
0&{otherwise}
\end{array}} \right.\]
where $\pi(d) $ is a normalization function and $\lambda_{i,c_j}$ is the weight of $f_j$ in maximum entropy $c_j$.The other parameters are defined in the same way as Section 3.2.1. After fifteen iterations of the improved iterative scaling algorithm \cite{pietra1997inducing} implemented in Natural Language Toolkit \cite{bird2006nltk}, the parameters of $\lambda _{i,c_j} $ are adjusted to maximize the entropy of distribution of training data.    
 
\subsubsection{Support Vector Machines} 
Support Vector Machines (SVM) is a very effective machine learning method firstly introduced by \cite{cortes1995support}. SVM constructs a hyperplane or a set of hyperplanes in a high dimensional space represented by $\vec{w}$. Since the larger the margin, the lower the error of the classifier, after training, the largest distance of support vector to nearest training-data point in any classes is achieved.
Then the problem of maximizing the margin turns to \[ \mathop {argmin\;}\limits_{\vec{w},\;b} \frac{1}{2}||w|{|^2}\]
where \[{y_i}(\vec{w} \cdot {x_i} - b) \ge 1\]
and its unconstrained dual form is the following optimization problem:
maximize $\tilde{L}(\mathbf{\alpha})$ where 
\begin{equation*}
\begin{aligned}
\tilde L({\bf{\alpha }}) & = \sum\limits_{i = 1}^n {{\alpha _i}}  - \frac{1}{2}\sum\limits_{i,j} {{\alpha _i}} {\alpha _j}{y_i}{y_j}k({{\bf{x}}_i},{{\bf{x}}_j})\\&= \sum\limits_{i = 1}^n {{\alpha _i}}  - \frac{1}{2}\sum\limits_{i,j} {{\alpha _i}} {\alpha _j}{y_i}{y_j}{\bf{x}}_i^T{{\bf{x}}_j}
\end{aligned}
\end{equation*}
subject to $\alpha_i \geq 0\ (i = 1, \ldots ,n)$.
Usually, the kernel here is linear, which means 
\[k(\mathbf{x}_i,\mathbf{x}_j)=\mathbf{x}_i\cdot\mathbf{x}_j\]

For SVM models, python tool scikit-learn   \cite{pedregosa2011scikit} is chosen for training and testing.
Scikit-learn\footnote{http://scikit-learn.org} was started in 2007 as a Google Summer of Code project, and has became the most efficient 
and useful tool for data mining and analysis in Python. With all default parameters, LinearSVC and SVC with linear kernel are used in our article.  
\subsection{Ensemble methods}

Ensemble methods \cite{dietterich2000ensemble,friedman2001greedy,ridgeway2007generalized} are supervised learning algorithm which commonly combine multiple hypotheses to form a better one. There are two families of ensemble methods,
averaging methods and boosting methods. In averaging methods, several estimators will be built to average their predictions. It is a kind of vote, namely, on average. The combined estimator is usually better than any of the fundamental estimators since its variance is reduced (e.g., Bagging methods and Forests of randomized trees). By contrast, in boosting methods, fundamental estimators are built sequentially and each one tries to reduce the bias of the combined estimator. The idea behind it is to combine several weak models to generate a more powerful ensemble model (e.g., AdaBoost and Gradient Tree Boosting).

The ensemble method modules are chosen from scikit-learn, including AdaBoost, Gradient Tree Boosting and Random Forests. For each Chinese review, the average value of word embeddings is used as the input. 

\subsection{CNN methods}
CNN is short for Convolutional Neural Networks. Its key module is to calculates the convolution between input and output. Just as CNN used in computer vision, a matrix is needed, as the input of CNN. After several experiments, we set $D=60$ as the dimension quantity of word embeddings for CNN. If there are $L$ words in a sentence, combine their word embeddings together to construct a matrix of size $L \times D$ as shown in Figure 1. $L=60$ is set since fixed L is needed, and which means, only 60 words are preserved from the beginning of a review. On the other hand, if the length is less than 60, the matrix will be filled with used vectors from the beginning of the review by repeating them. At last, every review is represented by a matrix of size $60 \times 60$. 
\begin{figure}[h]
  \centering
  %% insert PDF file testpdf.pdf
  \includegraphics[width=0.5\textwidth]{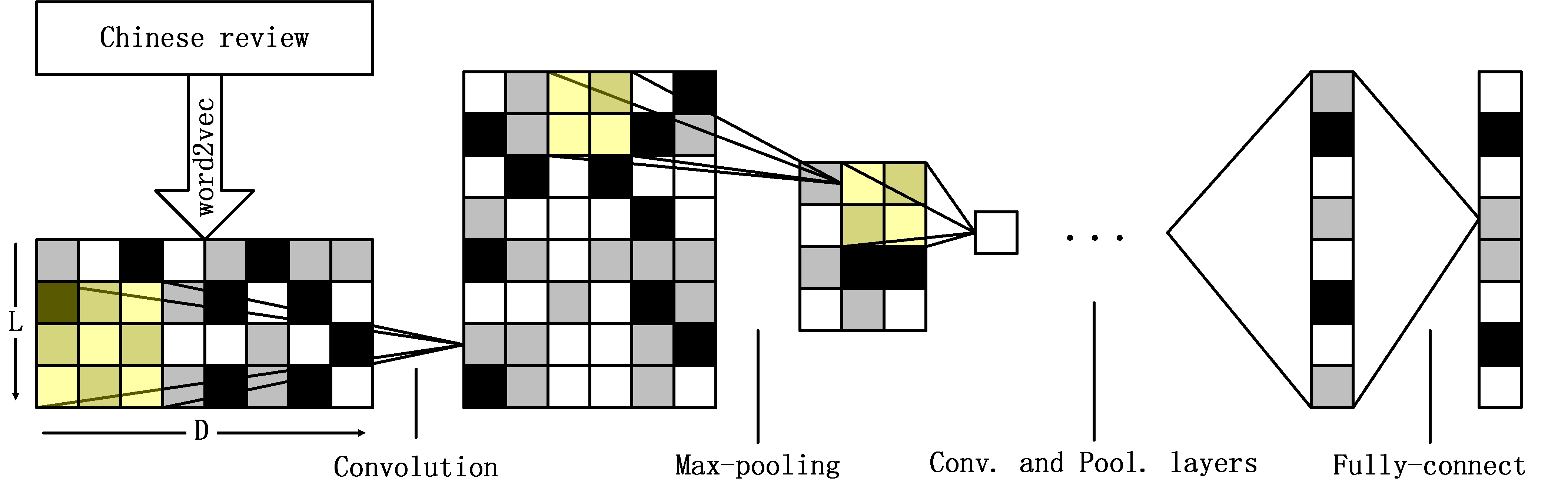}
  \caption{Illustration of Convnets}
  \label{fig:cummulative}
\end{figure}

Formally, in computer vision, given n images ($ X_l,\ l=1,...,n$) of size $r \times c$, k kernels of size $a\times b$ are set. For each kernel patches a small image($ X_s$) in the large image ($ X_l $),  $K(kernel_i,X_s)$ is computed, where $K()$ is the kernel function, giving us $k \times (r - a + 1) \times (c - b +1)$ array of convolved features (more detail,see  the tutorial\footnote{http://ufldl.stanford.edu/wiki/index.php}).
Max-pooling is the key module to help training deeper model. It works like this: Expect that there is a $60 \times 60$ matrix. Let's set pooling size $10 \times 10$, then the $60 \times 60$ matrix will be divided into 36 small matrixes of size $10 \times 10$. Just pick the biggest value in each small matrix and combine them together. At last a $6 \times 6$ matrix instead of $60 \times 60$ matrix is gotten. Extending the implementation \footnote{http://deeplearning.net/tutorial/lenet.html\#lenet} of the lenet5 \cite{lecun1998gradient}, the convolutional layer and max-pooling layer are merged as one layer. The structure of ConvNets used is shown in Table 2. 

\begin{table}[!hbp]
\begin{tabular}{ccccc}
\hline
Layer & Frame & Kernel &Kernel\_size & Pool \\
\hline
1& $60\times60$ & 40 &5$\times$5& 2$\times$1 \\
2& 28$\times$56 & 50 &5$\times$5& 2$\times$1 \\
3& 12$\times$52 & 50 &5$\times$5& 2$\times$1 \\
4& 4$\times$48 & $ - $ & $ - $ & $ - $ \\
\hline
\end{tabular}
\caption{Parameters of CNN layers }
\end{table}
With the fourth layer, a fully-connect sigmoidal layer is constructed to classify the output values. After experiments, there are some rules can be concluded:
\begin{figure*}
\centering
\subfigure[The worst performance of LR is worse than the best performance of NB]{ \label{fig:subfig:a} %% label for first subfigure
\includegraphics[width=3in]{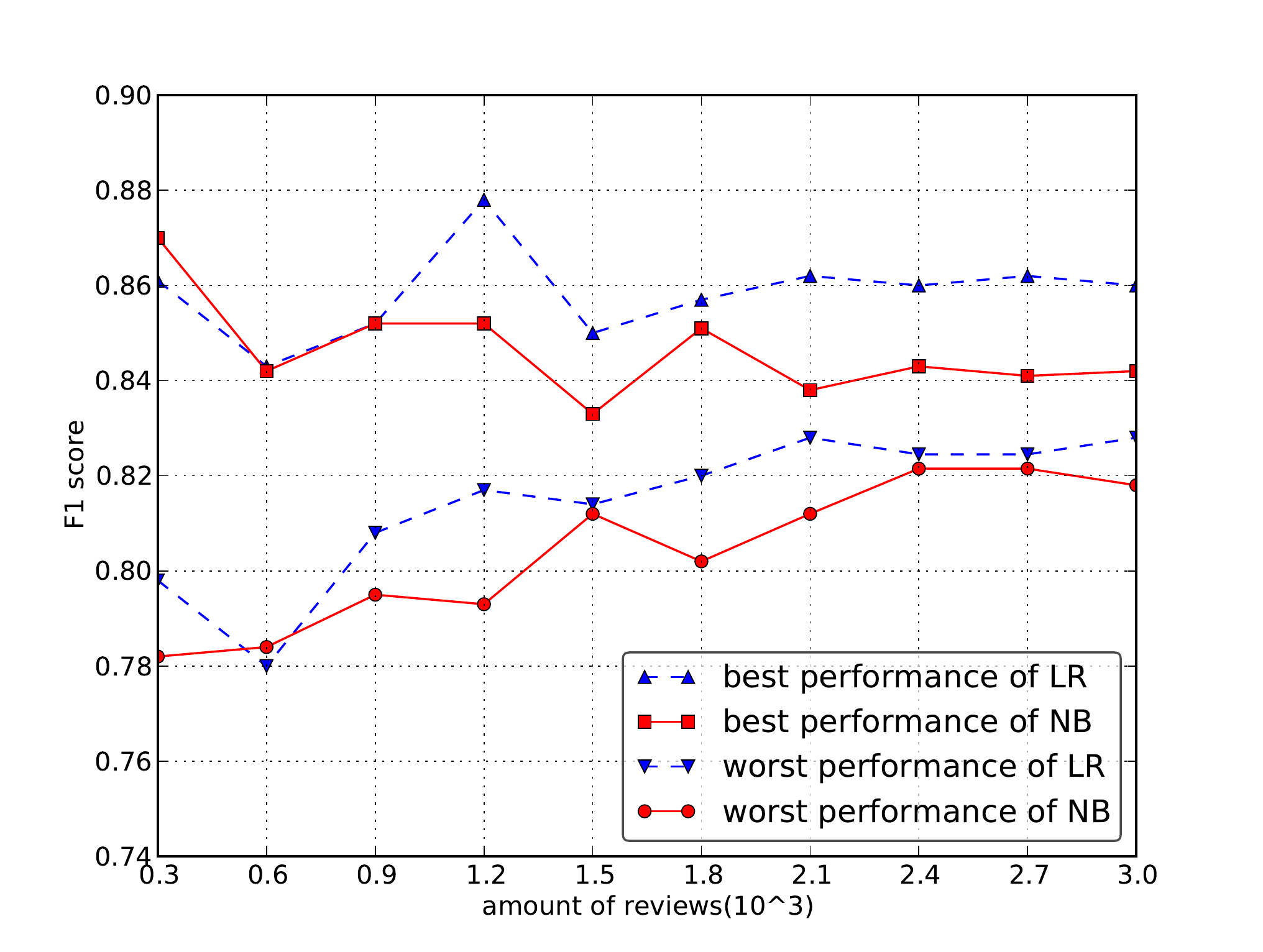}}
\hspace{0.31in}
\subfigure[The worst performance of LR is better than the best performance of NB]{ 
\label{fig:subfig:b} %% label for second subfigure 
\includegraphics[width=3in]{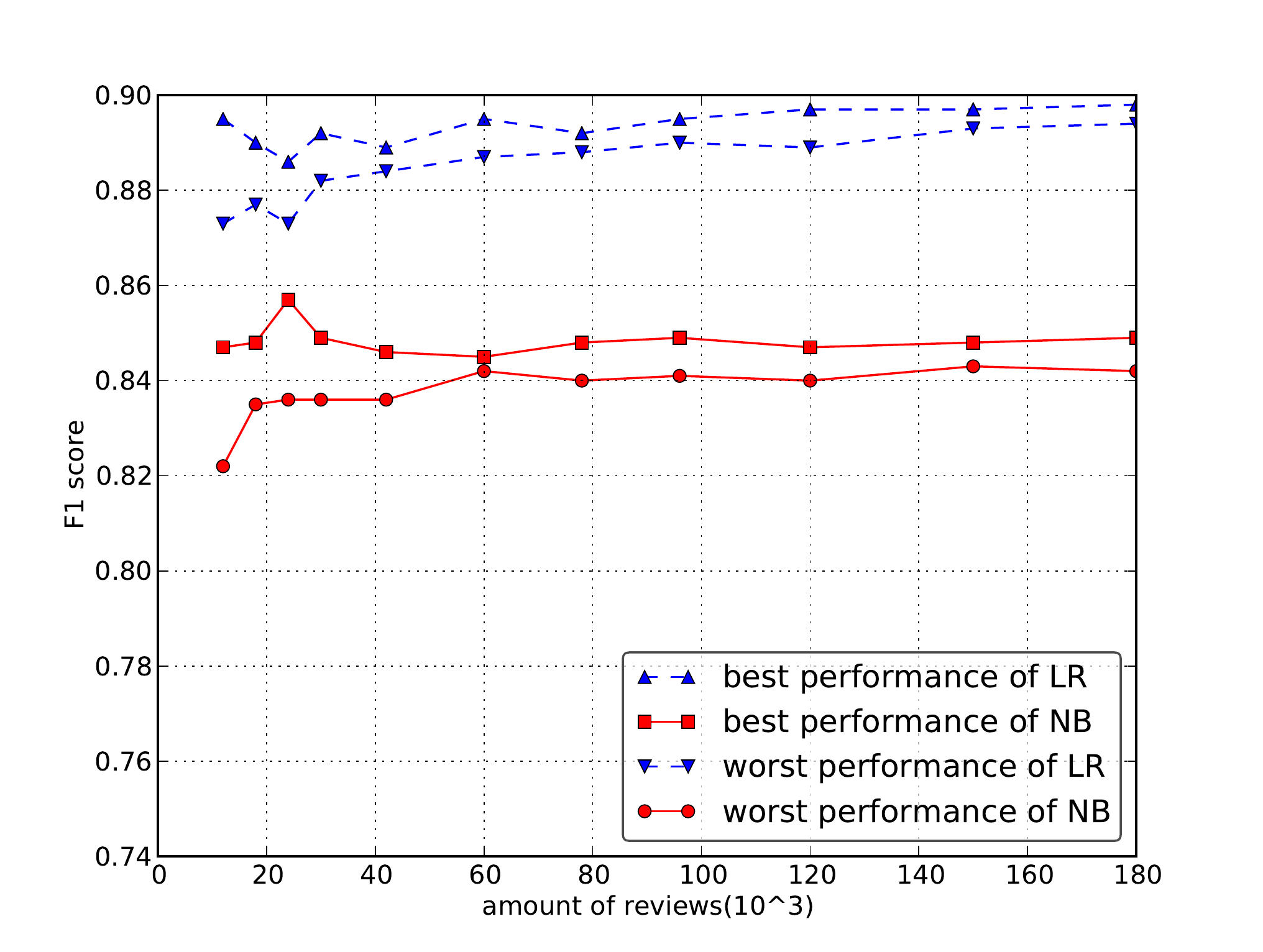}} 
\caption{The performance curves with different amount of reviews.}
\label{fig:subfig} %% label for entirefigure 

\end{figure*}
\begin{itemize}
\item The quantity of the word embedding dimensions shall be more than 50.
\item Do not use pooling between the dimensions of word embedding (thus, in Table 2, the size of pooling is 2$\times$1).
\item Adding more fully-connect sigmoidal layer dose not help in improving F1 score.
\end{itemize}

\section{Experiment results}
\subsection{Corpuses}
 Unlike English corpuses, Chinese corpuses are relatively small and usually focus on POS tagging \cite{mingqin2003building}, parsing \cite{xue2005penn} and translating \cite{xiao2010different}. In Chinese sentiment classification, the most 
popular corpus is ChnSentiCorp \cite{tan2008empirical} with 7,000 positive reviews and 3,000 negative reviews\footnote{http://www.datatang.com/data/11936}. Since the amount of data collected by previous Chinese NLP researchers is too small for our work, we build a new corpus, MioChnCorp, with a million Chinese hotel reviews. The corpus is public and can be downloaded directly\footnote{http://pan.baidu.com/s/1dDo9s8h}. The reviews are crawled and filtrated from the website\footnote{http://www.dianping.com/hotel} which has coarse-grained rating (5-star scale) for each review. We give up the 3-star reviews which may be ambiguous, and mark the five-star and four-star reviews as positive and the rest as negative. Finally 908189 positive reviews and 101762 negative reviews are obtained. After word segmentation\footnote{https://github.com/fxsjy/jieba } being done, the sentiment classification process is executed.

Since ChnSentiCorp is small, the result may be unstable. Thus, \newcite{tan2008empirical} gave the best performance and mean performance to evaluate a classification method. \newcite{zhai2011exploiting} tried to get a believable result using the average value from 30 experiments. See Figure 2, Naive Bayes and Logistic Regression are used as classification methods to show the performance curves with different amount of reviews. The first sub-graph is tested on ChnSentiCorp, the second on is tested on MioChNCorp. Balanced corpuses are split into 3 equal-sized folds, two for training, the rest for testing. After repeating each experiment five times, the best performance and worst performance are marked. At last, two conclusions can be made: Firstly, when the amount of reviews is less than 60,000, the performance will be improbable (the best performance of model minus worst performance is bigger than 0.01). Secondly, more data usually help to get better performance, but the performance will be finally stable when data are big enough (e.g., 120,000 reviews).   
\subsection{The performance measure}
F1 score (also called F-measure) is a measurement of a test's accuracy which combines recall and precision as follows:
\[{F_1} = 2 \cdot \frac{{Precision \cdot Recall}}{{Precision + Recall}}\]
\[{{Recall}} = \frac{{{\rm{correct\; positive\; predictions\;amount}}}}
{{{\rm{positive\; example\; amount}}}}\]
\[{Precision} = \frac{{{\rm{ correct\; positive\; predictions\; amount\;}}}}
{{{\rm{positive\; predictions\; amount}}}}\]
since there are two categories (positive and negative) in MioChnCorp, Macro F1 is used to  evaluate the performance of classification method over the corpus
\[Macro\;F1 = \frac{{Postive\;F1  + Negative\;F1}}{2}\]
in the rest of this article, F1 score means Macro F1 score. 

\subsection{Experimental design}
Nine methods are designed to classify  MioChnCorp using different features. NB and ME use 10543 words (the sentimental words and CHI words). LinearSVC use unigram and bigram. Five methods (SVC, LR, AdaBoost, Gradient Tree Boosting (GBT) and Random Forests (RF)) use the average vectors of word embeddings and CHI words  (extending the dimension quantity to 450). CNN use the matrix constructed by word embeddings from words in a review as feature.

Though all of these models are effective, the combination of different machine learning methods is supposed to acquire better F1 score. 
There are two ways to combine those methods. First is vote, the idea is simple, ``the minority is subordinate to the majority" (marked as Vote\_all ). The other way is to over-fit in the validate set. Add one more fold for validating into these tree folds. After training, nine models will be constructed. And each model gives one predication list for validating set. For each review, there are nine predications (e.g., [0 0 1 0 1 0 0 1 0 ], 0 means negative, 1 means positive). Using the predication vectors of validating reviews as input, and the labels of validating reviews as the  Logistic Regression output, after training on validating set, the combination model (called LR\_all) is built to test on testing set. The Framework of LR\_all is shown in Algorithm 1.  
\begin{algorithm}[htb] 
\caption{ Framework of combination model} 
\label{alg:Framwork} 
\begin{algorithmic}[1] %这个1 表示每一行都显示数字
\REQUIRE ~~\\ %算法的输入参数：Input
The experiment set of labelled samples:
\\$train\_{set}$, $validate\_{set}$ and $test\_set$;\\
$n$ machine learning classifiers (marked\\ as $C_i,\ i=1,...,n$) with default parameters 
\ENSURE ~~\\ %算法的输出：Output
For each classifier $C_i$, train  on $train\_set$;
\STATE Test on $validate\_set$ by $C_i$ and storing predication as $validate\_list\_i$;
\label{ code:fram:extract }%对此行的标记，方便在文中引用算法的某个步骤
\STATE Use logistic regression to predicate the label list of $validate\_set$ by combination data of {$validate\_list\_i$} ($i=1,...,n$) and store the model as $LR\_all$; 
\label{code:fram:trainbase}
\STATE For each classifier $C_i$, test over test set, and storing the predication as $test\_list\_i$; 
\label{code:fram:add}
\STATE Use $test\_list\_i$ as input of $LR\_all$ and produce final predication of $test\_set$, storing as $test\_list$
\label{code:fram:select}
\RETURN $C_i\ (i=1,...,n),\ LR\_all,\ test\_list$; %算法的返回值
\end{algorithmic}
\end{algorithm}

\subsection{Comparison and analysis}
Table 3, Table 4 and Table 5 show the performance of different machine learning methods. Subjected to hardware recourse (RAM:8G, CPU:Intel I5, GPU:GTX960M), the experiments are explored over corpuses with tree size: 40,000, 80,000 and 120,000. Each corpus is divided into four folds which are equal in size, two for training, one for validating, the rest for testing.   
\begin{table}[h]
\resizebox{0.5\textwidth}{!}{ 
\begin{tabular}{cccccc}
\hline
   & Pre\_0 &Rec\_0 & Pre\_1 & Rec\_1& F1 \\
\hline
NB& .843 & .896 &.889& .833& .864 \\
ME& .914 & .850 &.859& .910& .880 \\
LinearSVC& .898 & .881 &.883& .900 &.890 \\
LR& .902 & .879 & .882 & .905 &.892 \\
SVC& .910 & .878 & .882 & .913&.895 \\
Adaboost& .898 & .867 & .871 & .902 &.885 \\
GBT& .897 & .866 &.870& .890& .883 \\
RF& .910 & .850 &.860& .916& .883 \\
CNN& .905 & .865 &.870& .909& .887 \\
Vote\_all& .790 & \textbf{.955} &\textbf{.943}& .747 & .849 \\
LR\_all& \textbf{.915} & .893 &.896& \textbf{.917}& \textbf{.905} \\

\hline
\end{tabular}
}
\caption{Different performance over 40,000 reviews }
\end{table}

\begin{table}[h]
\resizebox{0.5\textwidth}{!}{
\begin{tabular}{cccccc}
\hline
   & Pre\_0 &Rec\_0 & Pre\_1 & Rec\_1& F1 \\
\hline
NB& .836 & .900 &.892& .823& .862 \\
ME& .907 & .853 &.861& .913& .883 \\
LinearSVC& .895 & .886 &.887& .897 &.891 \\
LR& .910 & .876 & .880 & .913 &.894 \\
SVC& .910 & .876 & .880 & \textbf{.914}&.895 \\
Adaboost& .899 & .866 & .871 & .903&.884 \\
GBT& .897 & .868 &.872& .901& .885 \\
RF& .910 & .868 &.874& \textbf{.914}& .891 \\
CNN& .904 & .864 &.869& .909& .886 \\
Vote\_all& .786 & \textbf{.957} &\textbf{.945}& .739 & .846 \\
LR\_all& \textbf{.915} & .895 &.897& .912& \textbf{.906} \\

\hline
\end{tabular}
}
\caption{Different performance over 80,000 reviews }
\end{table}

\begin{table}[!h]
\resizebox{0.5\textwidth}{!}{
\begin{tabular}{cccccc}
\hline
   & Pre\_0 &Rec\_0 & Pre\_1 & Rec\_1& F1 \\
\hline
NB& .839 & .900 &.892& .827& .863 \\
ME& .908 & .850 &.859& .913& .882 \\
LinearSVC& .900 & .891 &.892& .901 &.896 \\
LR& .897 & .882 & .884 & .900 &.890 \\
SVC& .905 & .881 & .884 & .907&.894 \\
Adaboost& .896 & .864 & .869 & .899&.882 \\
GBT& .896 & .867 &.871& .890& .883 \\
RF& .910 & .870 &.876& .914& .892 \\
CNN& .915 & .853 &.862& .920& .887 \\
Vote\_all& .777 & \textbf{.965} &\textbf{.953}& .724 & .842 \\
LR\_all& \textbf{.917} & .901 &.903& \textbf{.919}& \textbf{.910} \\

\hline
\end{tabular}}
\caption{Different performance over 120,000 reviews }
\end{table}
There are nine methods to construct the LR\_all model, but not all of them make contribution. Weka Explorer\footnote{http://www.cs.waikato.ac.nz/ml/weka/} provides attribute selection module to choose most useful attributes to the target attribute (namely, the label list of $validate\_set$ in our situation). Extracting the {$validate\_list\_i$} ($0<=i<n$) used in Algorithm 1, and combining these nine prediction lists with the label list of $validate\_set$, totally, ten attributes will be gotten.  With 10-fold cross-validation, CfsSubsetEval attribute evaluator and BestFisrt search Method, Weka selects five most valuable attributes (ME, SVC, LinearSVC, RF and CNN). It is reasonable because they are most outstanding machine learning models which represent their own feature selection methods. Considering the limit of hardware resource and running time, LR is used to instead of SVC and CNN is abandoned. The result is shown in Figure 3. To our surprise, even only four feature is chosen, the F1 score is not reduced. 
\begin{figure}[h]
  \centering
  %% insert PDF file testpdf.pdf
  \includegraphics[width=0.5\textwidth]{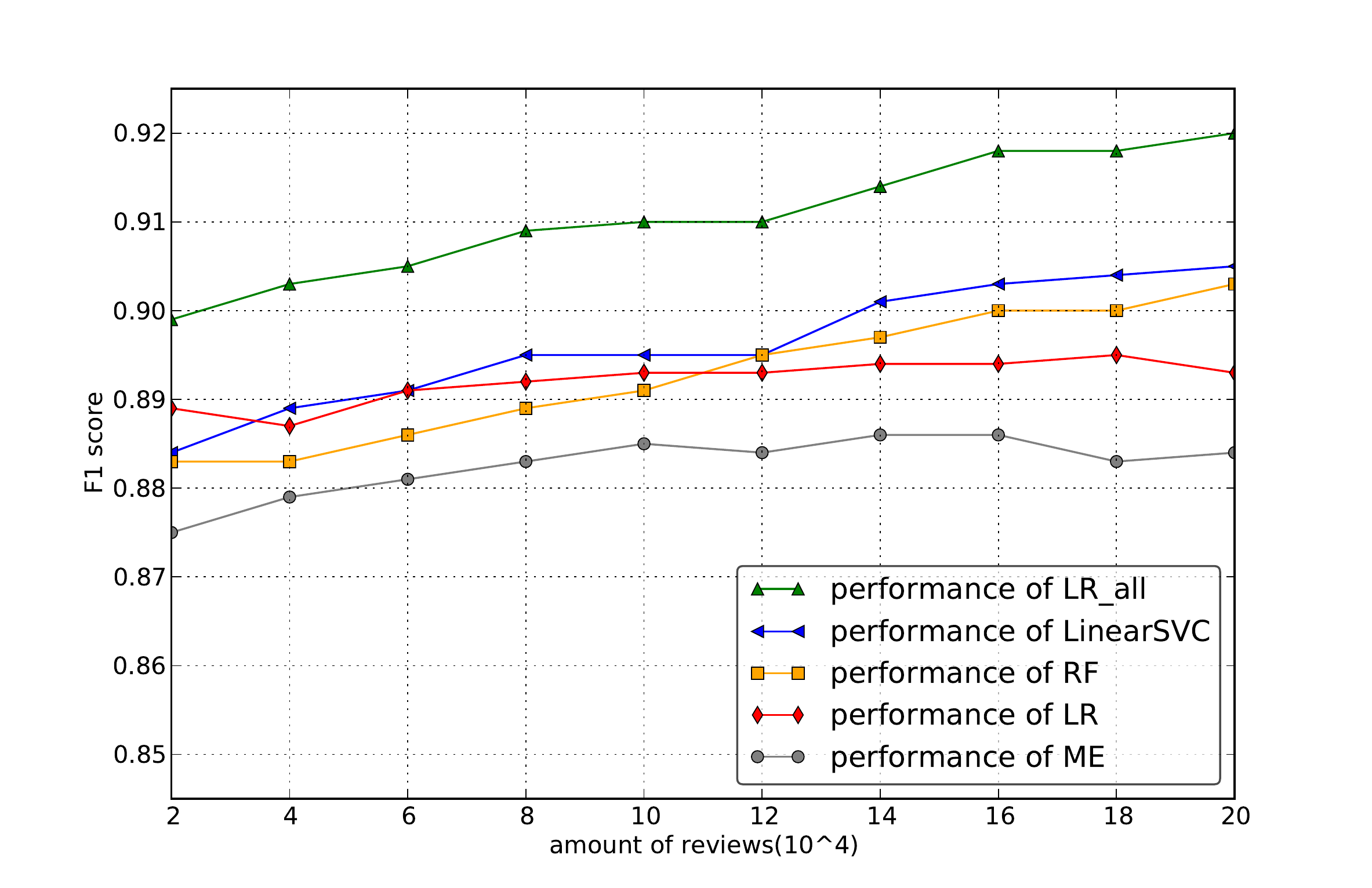}
  \caption{The performance curves of combination model and sub-models with different amount of reviews}
  \label{fig:cummulative}
\end{figure}

The more reviews we use in model building, typically the better performance we get till the performance is stable. SVM (linearSVC and SVC with linear kernel) has best performance not only in traditional bag of words models, but also in word embedding models. Three ensemble methods work similarly and bigger data help to improve their performance obviously. There may be three reasons why CNN works better than NB and ME, but does not reach the expectant performance. Firstly, the amount of reviews is not big enough to train a deep learning model. Secondly, the architecture of the model may not be enough suitable as a language model. Finally, the features (word embeddings with 60 dimensions) for CNN is not accurate enough to present syntax and semantics in sentence. Vote\_all does not work well in improving performance, but has the highest negative recall and positive precision. LR\_all has better performance than Vote\_all because the same weights chosen by Vote\_all make these sub-models are equally important.

\section{Conclusion and Future Work}
\label{sec:length}

In this article, an empirical study of sentiment
categorization on Chinese hotel review is introduced. In order to conduct this experiment, a Chinese corpus, MioChnCorp\footnote{http://pan.baidu.com/s/1dDo9s8h}, with a million Chinese hotel reviews is collected. Using MioChnCorp, a word2vec model is trained to present distributed representations of words and phrases in Chinese hotel domain.

Then the experimental results indicate that the more data we use, the better performance we get. And 60,000 or
larger size (e.g. 120,000) of reviews are sufficient in sentiment analysis of Chinese hotel review.

What's more, we employ word embeddings as input features without any sentiment lexicons, and find such features perform well by using ensemble methods, LR, SVM and CNN. With respect to these learning methods, SVM works best. Though CNN works not as good as expect, it still has better performance than NB and ME. The roles we used to construct the CNN model is introduce in Section 3.    

Finally, a methodology, LR\_all is constructed to combine different machine learning methods and get an outstanding performance in precision, recall and F1 score of 0.920. 

In the future, more work will be explored in building better CNN model for Chinese sentimental analysis and constructing combinational model in other tasks of NLP using word embedding.

\section{Acknowledgements}
The financial support for this work is provided by The Fundamental Research Funds for the Central Universities, No. ZYGX2014J065.

\bibliographystyle{acl}
\bibliography{submit_36}

\end{document}